\documentclass{article} % For LaTeX2e
\usepackage{latex_template,times}
\usepackage{hyperref}
\usepackage{url}
\usepackage{amsmath}
\usepackage{float}
\usepackage{amsfonts}
\usepackage{amssymb}
\usepackage{caption}
\usepackage{graphicx}
\usepackage{booktabs}
\usepackage{subcaption}

\title{Best Transition Matrix Esitimation or Best Label Noise Robustness Classifier? Two Possible Methods to Enhance the Performance of T-revision}

\nipsfinalcopy % Uncomment for camera-ready version

\begin{document}

\author{%
  Haixu Liu \\
  The University of Sydney\\
  \texttt{hliu2490@uni.sydney.edu.au} \\
  \And
  Zerui Tao \\
  The University of Sydney\\
  \texttt{ztao0063@uni.sydney.edu.au} \\
  \AND
  Naihui Zhang \\
  The University of Sydney\\
  \texttt{nzha3727@uni.sydney.edu.au} \\
  \And
  Sixing Liu \\
  The University of Sydney\\
  \texttt{sliu2285@uni.sydney.edu.au} \\
}

\maketitle

\begin{abstract}
Label noise refers to incorrect labels in a dataset caused by human errors or collection defects, which is common in real-world applications and can significantly reduce the accuracy of models. This report explores how to estimate noise transition matrices and construct deep learning classifiers that are robust against label noise.
In cases where the transition matrix is known, we apply forward correction and importance reweighting methods to correct the impact of label noise using the transition matrix. When the transition matrix is unknown or inaccurate, we use the anchor point assumption and T-Revision series methods to estimate or correct the noise matrix. In this study, we further improved the T-Revision method by developing T-Revision-Alpha and T-Revision-Softmax to enhance stability and robustness. Additionally, we designed and implemented two baseline classifiers, a Multi-Layer Perceptron (MLP) and ResNet-18, based on the cross-entropy loss function. We compared the performance of these methods on predicting clean labels and estimating transition matrices using the FashionMINIST dataset with known noise transition matrices. For the CIFAR-10 dataset, where the noise transition matrix is unknown, we estimated the noise matrix and evaluated the ability of the methods to predict clean labels.
Chapter 1 introduces the definition and impact of label noise, as well as the importance of our research, and Chapter 2 reviews related research work. In Chapter 3, we provide a detailed description of the baseline classifier architecture, loss function, and optimization strategies.
In Chapter 4, we present the motivation, principles, and implementation of the noise-robust loss functions for known label noise transition matrices, followed by comparative experiments and analysis. Chapter 5 covers the two transition matrix estimation methods and their improvements, along with experiments demonstrating the superiority of our improvements. Chapter 6 illustrates the complete training process for noise-robust classifiers in cases of unknown label noise and includes comparative experiments that further highlight the value of our work. The final chapter summarizes the experimental results, affirms the effectiveness of the T-Revision method, and discusses future work, including further optimization of transition matrix estimation and examination of the impact of matrix symmetry on noise estimation. The link to our code is available at \href{https://colab.research.google.com/drive/1f0SGvbRBNp4Jy1TOy6aFe38tcqnk3mKz?usp=sharing}{\textcolor{pink}{Colab}}, 
and the results can be found at \href{https://drive.google.com/file/d/19_ui2U6TvpzguuhgmOJegvJKDJX0LnYS/view?usp=sharing}{\textcolor{pink}{Results File}}.

\end{abstract}

\section{Introduction}
\label{intro}

With the development of neural networks, practitioners have increasingly recognized the importance of large-scale, high-quality labeled data. However, in real-world model training environments, obtaining fully reliable labels can be costly, and label noise is almost inevitable.
Label noise refers to the inaccurate labeling of samples in a dataset, which cannot fully reflect the true situation. \cite{frenay2013classification} It is usually caused by human errors in the data annotation process, inconsistent subjective cognition, defects in data collection equipment or processes, and noise class posterior. \cite{xiao2012adversarial} Label noise often degrades the performance of machine learning models, particularly in terms of accuracy and generalization ability, increase the demand for training data and model complexity, and change the frequency distribution of categories as the mislabeled sample will mislead the learning process. \cite{frenay2014comprehensive} In practical tasks, it can sometimes lead to serious consequences, such as misdiagnosis of diseases in the medical field, and safety threats due to misidentification of road conditions in the field of autonomous driving. Thus, addressing label noise is crucial in sensitive fields and learning from datasets containing label noises better has become increasingly significant, particularly in large-scale applications where manually verifying each label is infeasible. \cite{song2022learning}

A fundamental method for modelling the relationship between noisy labels and clean labels is to assume that the noise is independent of the features and depends only on the true labels, allowing a transition matrix to represent the conversion process between the true labels and noisy labels.Therefore, the transition matrix is \( T \in [0,1]^{n \times n}\), where \( T_{ij} = p( \bar{y} = j \mid y = i)  \). \cite{song2022learning} With the transition matrix, the output can be modified to match the true distribution of the true label, thus providing a more reliable loss and training process.

The aim of this project is to explore noise-robust classifiers through two different strategies: (1) leveraging known transition matrices and (2) estimating unknown noise transitions. We use three datasets provided by teaching team of Course COMP5328 at USYD, each featuring different noise characteristics. Two of them come with known class-conditional label noise rates, while the last one does not include a transition matrix, which requires us to estimate accurately. This project will implement two distinct classification models - Multi-Layer Perceptron (MLP) and ResNet -  and analyze their robustness against label noise by examining their performance on noisy validation data and clean test data. More specifically, we will use the given transition matrices from assignment introduction on FashionMINIST0.3 and FashionMINIST0.6 dataset to build noise-robust classifiers. For the CIFAR-10 dataset, we will develop a custom transition matrix estimator, validating its accuracy by comparing it with known matrices in the first two datasets. Additionally, we conducted multiple random sampling runs on the designed classifier models to produce stable and reliable evaluation results of the effectiveness and robusness.

\section{Previous Work}
Through our research, we found that representative studies on learning with noisy labels primarily come from the University of Sydney. We have selected some pioneering and representative works:

\subsection*{1. \textit{Making Deep Neural Networks Robust to Label Noise: A Loss Correction Approach \cite{patrini2017making}}}

\textbf{Motivation}: Deep neural networks are susceptible to performance degradation when trained with noisy labels. The goal of this work is to make networks robust to noisy labels by correcting the loss function.

\textbf{Method Overview}: This paper, serving as a baseline for this task, introduces two loss correction methods: Backward and Forward correction.
\begin{itemize}
    \item \textbf{Backward Correction}: This method compensates for the noise impact by correcting the loss function. Assuming there is a noise transition matrix \( T \) representing the probability of label transitions between classes, the loss is corrected by multiplying by the inverse matrix \( T^{-1} \). This ensures consistency with the "clean" label scenario, even under noise.
    \item \textbf{Forward Correction}: Unlike Backward correction, Forward correction adjusts the model's predictions by multiplying them with matrix \( T \), aiming to achieve the noisy label prediction distribution. Forward correction is mathematically straightforward but may rely on reasonable predictions for noisy labels.
\end{itemize}

\textbf{Advantages}:
\begin{itemize}
    \item \textbf{Solid Theoretical Foundation}: This method is based on rigorous theoretical derivations, ensuring classifier consistency under label noise.
    \item \textbf{General Applicability}: The loss correction method is agnostic to network structure and application domain, theoretically applicable to any deep network architecture.
    \item \textbf{Adaptability}: Both Forward and Backward correction methods provide strong noise robustness, with Forward correction showing better performance in various experiments.
\end{itemize}

\textbf{Disadvantages}:
\begin{itemize}
    \item \textbf{Dependency on Noise Matrix Estimation}: The effectiveness of this method heavily depends on the accurate estimation of the noise matrix. In some practical applications, estimating the noise matrix may be difficult or unstable, impacting effectiveness.
    \item \textbf{Difficulty in Matrix Inversion}: If the matrix is not full-rank, inversion can be challenging, limiting the Backward method's applicability.
\end{itemize}

\subsection*{2. \textit{Are Anchor Points Really Indispensable in Label-Noise Learning? \cite{xia2019anchor}}}

\textbf{Motivation}: Traditional label-noise learning methods rely on anchor points, which are difficult to obtain in complex datasets. This work aims to reduce dependency on anchor points.

\textbf{T-Revision Method}: The T-Revision method initializes the noise matrix through "approximate anchor points," identified as samples with high posterior probabilities for the noisy class. These samples serve as "approximate anchor points" to initialize the noise matrix. A relaxation variable is then added to dynamically adjust the noise matrix during training, optimizing it jointly with the classifier to approximate the true noise matrix.

\textbf{Advantages}:
\begin{itemize}
    \item \textbf{Reduced Anchor Point Dependency}: Traditional label-noise learning relies on anchor points, which are challenging to obtain. The T-Revision method introduces an alternative by using approximate anchor points, making it suitable for more practical scenarios.
    \item \textbf{Dynamic Noise Matrix Adjustment}: By adding a relaxation variable during training, the noise matrix can be continuously updated, enhancing model robustness against label noise.
\end{itemize}

\textbf{Disadvantages}:
\begin{itemize}
    \item \textbf{Dependency on Initialization}: This method relies on selecting initial approximate anchor points and high posterior probability samples. If these samples are inaccurate or unevenly distributed, the initialization of the noise matrix may be affected. Additionally, negative values in the initialization of the correction matrix combined with Importance Reweighting can lead to negative loss values, making training unstable.
    \item \textbf{High Computational Complexity}: The approach involves training an estimator, pretraining with the T matrix to estimate clean label probabilities, and continuously adjusting the noise matrix while fine-tuning the classifier, resulting in a significant computational burden.
\end{itemize}

\subsection*{3. \textit{Classification with Noisy Labels by Importance Reweighting \cite{liu2015classification}}}

\textbf{Motivation}: To mitigate the impact of label noise on model training by adjusting sample weights, ensuring classifier consistency.

\textbf{Importance Reweighting Method}:
Define the weight \( \beta(X, Y) \) as:
\begin{equation}
\beta(X, Y) = \frac{P_D(Y|X)}{P_{D_\rho}(Y|X)},
\end{equation}

This ratio of the clean conditional probability \( P_D(Y|X) \) to the noisy conditional probability \( P_{D_\rho}(Y|X) \) allows loss correction if we can accurately estimate the noisy label conditional probability. For each sample, the loss \( \ell(f(X), Y) \) is multiplied by the weight \( \beta(X, Y) \). This reweighting lets the model optimize with noise-adjusted data, where each noisy sample's loss is appropriately modified. To obtain reliable reweighting, the authors propose three methods to estimate noisy conditional probabilities:
\begin{itemize}
\item \textbf{Kernel Density Estimation (KDE)}: KDE is used to estimate the noisy conditional probability using Bayesian formula denominators. KDE is non-parametric and works well for low-dimensional data with large samples but struggles with convergence in high-dimensional data due to high computational costs.
\item \textbf{Density Ratio Estimation}: Directly estimates the density ratio \( r(X) = \frac{P_{D_\rho}(X|Y)}{P_{D_\rho}(X)} \), avoiding separate estimation of the numerator, reducing complexity in high dimensions. The paper uses Bregman divergence-based ratio matching to estimate noisy conditional probabilities. It is more efficient than KDE and suitable for high-dimensional data.
\item \textbf{Direct Probability Classification}: This method trains a classifier to directly estimate \( P_{D_\rho}(Y|X) \), mapping classifier outputs into \([0, 1]\) using link functions (e.g., sigmoid or softmax) to interpret them as probabilities. This approach, which converts clean conditional probabilities to noisy ones via transition matrices, is adopted in our later work.
\end{itemize}

\textbf{Advantages}:
\begin{itemize}
    \item \textbf{High Versatility}: This method can be applied to any surrogate loss function and is theoretically compatible with various classification algorithms, particularly useful for noisy labels.
    \item \textbf{Consistency Guarantee}: The authors theoretically prove that importance reweighting ensures classifier consistency under noise, making the final classifier similar to one trained on clean data.
\end{itemize}

\textbf{Disadvantages}:
\begin{itemize}
    \item \textbf{Complex Estimation Process}: Estimating conditional probabilities and noise rates is complex and may require extensive data. KDE is inefficient in high-dimensional data, and the process's computational cost is high due to density estimation and weight computation.
\end{itemize}

\subsection*{4. \textit{Dual T: Reducing Estimation Error for Transition Matrix in Label-noise Learning \cite{yao2020dual}}}

\textbf{Motivation}: Existing methods rely on estimating noisy class posterior probabilities, introducing significant error due to label noise randomness. This paper proposes a new estimation method by introducing intermediate classes to reduce dependency on noisy posterior probabilities.

\textbf{Dual-T Method}: This method introduces an intermediate class to decompose the original noise transition matrix into two sub-matrices, avoiding direct estimation of noisy posterior probabilities and reducing estimation bias caused by label noise.

\textbf{Advantages}:
\begin{itemize}
    \item \textbf{Reduced Estimation Error}: By decomposing the matrix, Dual-T avoids direct estimation of noisy posterior probabilities, significantly reducing error. Experimental results show superior classification performance over traditional label-noise learning methods across various datasets.
\end{itemize}

\textbf{Disadvantages}:
\begin{itemize}
    \item \textbf{Sample Size Limitation}: In cases with small sample sizes or uneven class distributions, the estimation error of the Dual-T method may be large.
\end{itemize}

\section{Implemented Baseline Classifier}
\label{3}
Since the main focus of this paper is to improve the loss function in classification tasks to enhance robustness to label noise, we implemented two classification models, ResNet-18 (CIFAR-10) and a Multi-Layer Perceptron (MLP), using the cross-entropy loss function as the baseline. This setup allows us to evaluate the performance of our reproduced and improved methods across different classifiers and datasets. For ResNet-18, we used the default hyperparameters, while for the MLP model, we designed a simple 5-fold cross-validation experiment to select hyperparameters and training strategies for both models. Given the limitations of article length and verification time, as well as the fact that this is not the main focus of our work, we only present the final selected model hyperparameters and training strategies below.

\subsection{Classification Model}

\subsubsection{Multi-Layer Perceptron (MLP)}
The Multi-Layer Perceptron (MLP) model used in this study is a fully connected neural network designed for 4-class classification. The MLP model takes a flattened input and processes it through three hidden layers with decreasing neuron counts: 2048, 1024, and 512. Moreover, each hidden layer is followed by a ReLU activation function to introduce non-linearity, a Batch Normalization layer to stabilize learning, and a Dropout layer with a dropout rate of 0.2 to prevent overfitting.

\textbf{Dropout} is a regularization technique used to prevent overfitting in neural networks by randomly “dropping out” a proportion of neurons during each training iteration. This process can force the model to learn more robust features by preventing any single neuron from becoming Over-dependent. Thus, the dropout technique can enhance generalization and improve the model’s ability to perform well on unseen data.

\textbf{Batch normalization} is also applied to stabilize and accelerate neural network training. For each mini-batch, it normalizes the inputs to mean=0 and variance=1. Importantly, BatchNorm can migrate the problem of internal covariate shift and reduce the effects of exploding and vanishing gradients.

While the MLP offers simplicity and computational efficiency, regularization techniques such as Dropout and Batch Normalization further enhance its generalization capabilities, making it robust. These make MLPs a more foundational and effective model for different datasets.

\subsubsection{ResNet-18}

ResNet employs a residual learning framework that uses residual blocks and skip connections to learn residual mappings.  According to \cite{he2016deep}, skip connections allow gradients to bypass certain layers, effectively mitigating the vanishing gradient problem in deep networks. This design improves the training stability of deep networks by enabling smoother gradient flow.

In the family of ResNet, the ResNet-18 (CIFAR-10) is implemented in this task.
\begin{itemize}
    \item \textbf{Initial Convolutional Layer}: A \(3 \times 3\) convolutional layer with an input channel of 3, output channel of 64, stride of 1, and padding of 1. Unlike the ImageNet version, this CIFAR-10 variant does not use a \(7 \times 7\) convolution layer.
    
    \item \textbf{Four Residual Block Groups}:
    \begin{enumerate}
        \item \textbf{First Group}: Two \(3 \times 3\) convolutional layers, with an output channel of 64 and stride of 1.
        \item \textbf{Second Group}: Two \(3 \times 3\) convolutional layers, with an output channel of 128. The first layer has a stride of 2 for downsampling.
        \item \textbf{Third Group}: Two \(3 \times 3\) convolutional layers, with an output channel of 256. The first layer has a stride of 2.
        \item \textbf{Fourth Group}: Two \(3 \times 3\) convolutional layers, with an output channel of 512. The first layer has a stride of 2.
    \end{enumerate}

    \item \textbf{Global Average Pooling Layer}: Applied after all residual blocks, reducing each channel’s feature map to a \(1 \times 1\) size.
    
    \item \textbf{Fully Connected Layer}: The final classification layer, with an output equal to the number of classes (4 for the our dataset).
\end{itemize}

Each block learns the residual function $F(x)$, with the output computed as 
$F(x)+x$. This design improves convergence efficiency without sacrificing accuracy, making ResNet-18 ideal for moderate-complexity tasks. It is also the baseline for the image classification task.
\subsection{Training Strategies}
\subsubsection{Optimization Method - Adam}
During the parameter optimization, the Adam (Adaptive Moment Estimation) optimizer is used, helping balance convergence speed with stability. This can ensure faster and more reliable training outcomes. Keeping the advantages of both the AdaGrad and RMSprop algorithms, Adam adapts the learning rate for each parameter individually by estimating the first and second moments of the gradients. \cite{kingma2014adam} This allows the optimizer to perform well on sparse gradients and noisy data, making it effective for deep neural networks. Adam's update rule is as follows:
\begin{align}
m_t &= \beta_1 m_{t-1} + (1 - \beta_1) g_t, \notag\\
v_t &= \beta_2 v_{t-1} + (1 - \beta_2) g_t^2 ,\notag\\
\hat{m}_t &= \frac{m_t}{1 - \beta_1^t}, \\
\hat{v}_t &= \frac{v_t}{1 - \beta_2^t}, \notag\\
\theta_{t+1} &= \theta_t - \frac{\alpha}{\sqrt{\hat{v}_t} + \epsilon} \hat{m}_t,\notag
\end{align}
where: \( \theta_{t+1} \) represents the updated parameters at step \( t+1 \); \( \alpha \) is the learning rate; \( \beta_1 \) and  \( \beta_2 \) are the exponential decay rate for the first and second moment estimates; \( g_t \) is the gradient of the objective function at step \( t \); \( \epsilon \) is a small constant to prevent division by zero; \( m_t \) and \(v_t \) are the first and second-moment estimate; \( \hat{m}_t \) and \( \hat{v}_t \) are the bias-corrected first and second-moment estimate.

\subsubsection{Loss Function}
The Cross-Entropy (CE) loss function is commonly used in multi-label classification tasks, particularly in deep learning models. It measures the difference between the predicted probability distribution and the true distribution. The Cross-Entropy loss evaluates each label independently and sums the losses across all labels. The formula is defined as:

\begin{equation}
\ell = - \sum_{i=1}^{C} \left[ y_i \cdot \log(p_i) + (1 - y_i) \cdot \log(1 - p_i) \right]
\end{equation}

where \(C\) is the number of labels, \(y_i\) is the true label (0 or 1), and \(p_i\) is the predicted probability.

\subsubsection{Regularization for Avoiding Overfitting}
Early stopping is used to prevent overfitting during the training. \cite{prechelt2002early}
By monitoring the model’s validation performance, training is stopped when the validation loss is not improving for a set number of epochs. The early-stopping approach can help balance model performance with generalization. This can lead to improved performance on unseen examples while reducing computational costs by avoiding unnecessary training epochs.

\section{Label Noise Methods With Known Flip
Rates }
Suppose training a classification model using a noisy dataset \( \bar{\mathcal{S}} \) containing pairs \( (X, \bar{Y}) \), where \( X \) is an instance, and \( \bar{Y} \) is a noisy label (potentially correct or incorrect). The aim is to predict the clean label \( Y \) by fully leveraging this noisy data.

To handle label noise, we use a transition matrix \( T \) to contain the probability of observing a noisy class\( \bar{Y} \) given the true class \( Y \). In this paper, the \( T \) matrix is a row-stochastic matrix \cite{patrini2017making}.
Each element \( T_{ij} \) represents the probability of a sample with the true class \( i \) being flipped to a class \( j \). And the relationship between the noisy class posterior \( P(\bar{Y} = j | x) \) and the clean class posterior \( P(Y = i | x) \) can be represented as formula \ref{relationship}:

\begin{equation}
\label{relationship}
P(\bar{Y} = j | x) = \sum_{i=1}^{N} P(\bar{Y} = j | Y = i, x) P(Y = i | x)
\end{equation}

where \( N \) is the total number of classes. This approach accounts for label noise in training, which can improve generalization to the clean label distribution.
\subsection{Dataset Description and Preprocessing}
We use the grayscale image dataset FashionMINIST here to evaluate the performance of proposed algorithms. The dataset involves two slightly different replicas: FashionMINIST0.3 and FashionMINIST0.6, corresponding to different levels of label noise intensity and given transition matrices. 
These two datasets each contain 28000 samples of size \(28\times28\), of which 4000 are used for testing and the rest for training. Thus, during the preprocessing, the fisrt thing we did is resizing all images from the original 28×28 to a fixed size of 32×32, which matches the CIFAR-10 dataset we would use in Section \ref{unknown noise method}, ensuring no errors are thrown due to dimension issues when processed by ResNet18 (CIFAR-10). 
For these grayscale images, we keep a single-channel output and convert the images to tensor format to meet PyTorch requirements. Additionally, when training the MLP model, we flatten the image tensor from three dimensions (channel, height, and width) into a single-dimensional tensor. It is worth noting that, since we did not use any pre-trained models, we did not normalize the pixel value range.

\subsection{Label Noise Robustness Loss Functions}
\subsubsection{Forward Correction}
Patrini et al. proposed an approach to train deep neural networks, subject to class-dependent label noise to mitigate the effects of noisy labels. Given the transition matrix \( T \) containing label flipping rates is known, it aims to modify a loss function \( \ell \) making it robust to asymmetric label noise. Given this assumption, it introduces a “forward correction” technique where the loss function is adjusted to account for label noise. \cite{patrini2017making} Moreover, it is agnostic to both application domain and network architecture.  More importantly, this method ensures that the model learns based on corrected signals, making it more robust to noise while avoiding numerical instability from the matrix inversion. \cite{patrini2017making}

In the forward correction, the noisy prediction \( \hat{p}(y|x) \) is adjusted by multiplying it with the transpose of transition matrix \( T \). This transition matrix \( T \) is applied to the output of the neural network before the loss calculation. Therefore, it can obtain a more aligned representation of the true label distribution and the adjusted  probability predictions become:

\begin{equation}
    \hat{p}_{\text{adjusted}}(y|x) = T^\top \hat{p}(y|x)
\end{equation}

To analyze the behavior of forward correction, we first need to consider a link function \(\psi : \Delta^{-1} \rightarrow \mathbb{R}^c\) that is invertible. 
\(\Delta^{c-1}\) is a set that contains all non-negative \(c\)-dimensional vectors whose elements sum to 1. In other words, it includes all possible probability distributions over \(c\) classes.

Formally, it is defined as:
\begin{equation}
\Delta^{c-1} = \left\{ p \in \mathbb{R}^c \,|\, p_i \geq 0 \text{ for all } i \text{ and } \sum_{i=1}^c p_i = 1 \right\},
\end{equation}

Many losses are said to be composite, denoted by \(\ell_{\psi} : \mathcal{Y} \times \mathbb{R}^c \rightarrow \mathbb{R}\), where  \(\ell_{\psi}\) represents a loss function that takes a true label \(y \in \mathcal{Y}\) and a prediction vector \(h(x) \in \mathbb{R}^c\) as inputs, and outputs a real number as the loss value, which measures the error between the prediction and the true label.
It can be expressed by the aid of a link function as:
\begin{equation}
\ell_{\psi}(y, h(x)) = \ell(y, \psi^{-1}(h(x))).
\end{equation}

In the case of cross-entropy, the softmax function serves as the inverse link function. where \( h(x) \) is the origin output from the neural network without softmax, \( \ell \) is the base loss function. Now we introduce the label noise.
Suppose that the noise matrix \(T\) is non-singular. Given a proper composite loss \(\ell_{\psi}\), define the forward loss correction as:
\begin{equation}
\ell_{\psi}^{\rightarrow} (h(x)) = \ell(T^\top \psi^{-1}(h(x))).
\end{equation}

In this task, original loss function cross-entropy is used.The implementation is such that we split the cross-entropy function into calculating softmax first and then the negative log-likelihood loss, so the actual change is that we corrected the output of softmax with a transfer matrix, thus the corrected loss function should become:

\begin{equation}
    \ell_{\text{forward}}(y, h(x)) = -\log( T^\top \hat{p}(y|x)) .
\end{equation}

Therefore, the forward correction technique offers an effective way to enhance the resilience of deep learning models to noisy labels, and the objective function is:
\begin{equation}
\arg \min_{h} \mathbb{E}_{x,y} \ell_{\text{forward}}(y, h(x))
\end{equation}

\subsubsection{Importance Reweighting}
The importance reweighting method is implemented to address the challenge of label noise. The traditional loss functions may fail to represent the true data distribution, leading to biased models. Importance reweighting corrects for this by rewriting the expected risk during training. This reweighting enables the model to focus on samples that more likely represent the true label distribution, reducing the impact of noisy labels. \cite{xia2019anchor} Importantly, it allows any surrogate loss function designed for traditional classification problems to be used effectively in noisy settings, as it maintains the convexity of the objective function and integrates well with batch learning optimization procedures. \cite{liu2015classification}

The importance reweighting method adjusts the empirical risk minimization process by weighting each sample based on the probability that its observed label reflects its true class. Following \cite{xia2019anchor}, the empirical risk $R_{\ell,D}(f)$ can be defined as:

\begin{equation}
R_{\ell,D}(f) = R\left[ D,f,\ell \right]= 
\mathbb{E}_{X,Y \sim D} \left[ \ell(f(X), Y) \right] = \mathbb{E}_{X,\bar{Y} \sim D_{\rho}} \left[ \beta(X, \bar{Y}) \ell(f(X), \bar{Y}) \right]
\end{equation}

\text{where $f$ denotes the classifier function; $D$ denotes the distribution for clean data,}
$D_{\rho}$ for noisy data; $\beta(X, \bar{Y}) = \frac{P_D(X, Y)}{P_{D_{\rho}}(X, \bar{Y})}$.

For the problem of classification in the presence of label noise, the label noise is assumed to be independent of instances as \( P_D(X) = P_{D_{\rho}}(X) \), therefore have:

\begin{equation}
\beta(X, \bar{Y}) = \frac{P_D(X, Y)}{P_{D_{\rho}}(X, \bar{Y})} = \frac{P_D(Y | X) P_D(X)}{P_{D_{\rho}}(\bar{Y} | X) P_{D_{\rho}}(X)} \\
= \frac{P_D(Y | X)}{P_{D_{\rho}}(\bar{Y} | X)}.
\end{equation}
where the conditional probability $P_{D_{\rho}}(\bar{Y} | X)$ can be estimated by a probabilistic classification method, and thus can be interpreted as probabilities. \cite{xia2019anchor} Therefore, the weight $\beta(X, \bar{Y})$ can be learned by using the noisy sample and the noise rates. 

Thus, if the transition matrix \( T \) is given, The loss will be modified through the following steps:

\begin{enumerate}
    \item \textbf{Noisy posterior probability}: The classifier predicts the posterior probability for the noisy label $\bar{Y}_i$, given the input $X = x$, denoted as:
    \begin{equation}
    P(\bar{Y}_i \mid X = x) = g_{\bar{Y}_i}(x),
    \end{equation}
    where $g(x)$ is the softmax output of the classifier.

    \item \textbf{Transformed posterior probability}: The posterior probability is then transformed by the transition matrix $T$ to approximate the noisy label distribution:
    \begin{equation}
    \hat{P}(\bar{Y}_i \mid X = x) = T^\top g(x).
    \end{equation}

    \item \textbf{Importance weighting}: The importance weight $\beta$ is calculated as the ratio of the clean and noisy posterior probabilities:
    \begin{equation}
    \beta = \frac{P(\bar{Y} \mid X = x)}{\hat{P}(\bar{Y} \mid X = x)} = \frac{g_{\bar{Y}}(x)}{(T^\top g(x))_{\bar{Y}}}.
    \end{equation}

    \item \textbf{Weighted cross-entropy loss}: The final loss is weighted using $\beta$ to account for the discrepancy between the noisy and clean labels:
    \begin{equation}
    \ell_{\text{weighted}}(f(x), \bar{Y}) = \sum_{i=1}^{n} \beta_i \cdot \ell(f(x), \bar{Y}_i).
    \label{eq:loss}
    \end{equation}
\end{enumerate}

\subsection{Experiment}

\subsubsection{Experiment Setting}
Since the transition matrices for these two datasets are provided, no additional estimation is required. We used the forward correction and importance reweighting methods with MLP and ResNet to predict the labels. To minimize the impact of randomness in training on the experimental results, we conducted ten experiments, each time randomly splitting the training data into training and validation sets in an 8:2 ratio. We calculated the mean of the results from these 10 experiments to compare model performance and computed the standard deviation to demonstrate the stability of our model.

\subsubsection{Evaluation Metrics}

The performance of each classifier will be evaluated using the top-1 accuracy metric. It is straightforward to calculate and interpret, which is defined as follows:

\begin{equation}
\text{Top-1 Accuracy} = \frac{\text{number of correctly classified examples}}{\text{total number of test examples}} \times 100\%
\end{equation}
This paper also uses the cross-entropy loss of the testset as an evaluation criterion.
\subsubsection{Experiment Results and Discussion}
 The results for FashionMINIST0.3 are in Table \ref{tab:MINIST0.3 RESULT}.
\begin{table}[h]
    \centering
    \begin{tabular}{lcccc}
        \toprule
        \textbf{Method} & \textbf{Loss Mean} & \textbf{Loss STD} & \textbf{Accuracy Mean (\%)} & \textbf{Accuracy STD} \\
        \midrule
        Baseline ResNet & 0.4536 & 0.0431 & 94.9600 & 0.4546 \\
        Forward ResNet & 0.1237 & 0.0054 & 96.1600 & 0.2035 \\
        Importance ResNet & 0.1329 & 0.0075 & 95.8150 & 0.1726 \\
\hline
        Baseline MLP & 0.5019 & 0.0124 & 93.1400 & 0.2555 \\
        Forward MLP & 0.2133 & 0.0221 & 93.5500 & 0.3693 \\
        Importance MLP & 0.2377 & 0.0377 & 93.5100 & 0.5100 \\
        \bottomrule
    \end{tabular}
\caption{Testset Performance Metrics for Different Models on FashionMINIST0.3}
\label{tab:MINIST0.3 RESULT}
\end{table}

By comparing the results of ResNet with different noise robustness methods, it is found that the Baseline model has a test loss mean of 0.4536 and an accuracy mean of 94.96\%. This model shows high loss and large standard deviation in the presence of label noise. After applying the Forward Correction method, the test loss significantly decreases to 0.1237, and accuracy improves to 96.16\%, with reduced loss and accuracy standard deviations. This indicates that Forward Correction effectively mitigates the impact of label noise, enhancing the model's stability and accuracy. 
Although the Importance Reweighting method achieves a test loss of 0.1329 and an accuracy of 95.8150\%, slightly lower than Forward Correction, it still performs better than the Baseline, showing some improvement. However, its standard deviations (0.0075 and 0.1726) remain relatively high, possibly due to inaccurate label confidence estimates that cause the model to overemphasize certain mislabeled samples. The overall test set accuracy and Loss performance of the MLP model is lower than that of the Resnet model, but the same conclusions can be drawn as for the ResNet

\begin{table}[h]
    \centering
    \begin{tabular}{lcccc}
        \toprule
        \textbf{Method} & \textbf{Loss Mean} & \textbf{Loss STD} & \textbf{Accuracy Mean (\%)} & \textbf{Accuracy STD} \\
        \midrule
        Baseline ResNet & 1.0134 & 0.0233 & 89.6525 & 1.6813 \\
        Forward ResNet & 0.3165 & 0.0449 & 90.4875 & 0.8798 \\
        Importance ResNet & 0.3423 & 0.0532 & 90.1850 & 1.1169 \\
        \hline
        Baseline MLP & 1.0326 & 0.0350 & 88.7550 & 1.1938 \\
        Forward MLP & 0.2964 & 0.0166 & 91.1050 & 0.5529 \\
        Importance MLP & 0.3272 & 0.0285 & 90.8150 & 0.8890 \\
        \bottomrule
    \end{tabular}
    \caption{Testset Performance Metrics for Different Models on FashionMINIST0.6}
    \label{tab:MINIST0.6 RESULT}
\end{table}
From Table \ref{tab:MINIST0.6 RESULT},
It can be observed that the Baseline ResNet has a mean test loss of 1.0134 and an accuracy mean of 89.6525\%, with a relatively high test loss and a large accuracy standard deviation (1.6813), indicating poor robustness in high noise (0.6) conditions. After applying the Forward Correction method, the test loss of ResNet significantly decreases to 0.3165, and the accuracy improves to 90.4875\%, with reduced loss and accuracy standard deviations of 0.0449 and 0.8798, respectively, indicating that the Forward Correction method effectively mitigates the impact of label noise, enhancing the model's stability and accuracy. The Importance Reweighting method achieves a test loss of 0.3423 and an accuracy of 90.1850\%, slightly lower than the Forward Correction method but still showing a significant improvement over the Baseline model. Its loss standard deviation is 0.0532, and the accuracy standard deviation is 1.1169, indicating that some label noise estimates may be inaccurate, causing the model to overemphasize certain mislabeled samples and affecting model stability.In comparison, the Baseline MLP has a mean test loss of 1.0326 and an accuracy mean of 88.7550\%, with loss and accuracy standard deviations of 0.0350 and 1.1938, indicating that the Baseline MLP exhibits poor stability under high label noise. After applying the Forward Correction method, the test loss of MLP significantly decreases to 0.2964, and the accuracy improves to 91.1050\%, with reduced loss and accuracy standard deviations of 0.0166 and 0.5529, respectively, showing better stability and accuracy. The Importance Reweighting method for MLP achieves a test loss of 0.3272 and an accuracy of 90.8150\%, slightly lower than the Forward Correction method but still better than the Baseline model, with loss and accuracy standard deviations of 0.0285 and 0.8890. This suggests that this method also brings some stability improvements to MLP, though its performance is slightly inferior to the Forward Correction method.

Additionally, by comparing the results of the same method on datasets with different noise levels, we can observe that the higher the noise rate, the poorer the model’s classification performance.

\section{Noise Rate Estimation Method}
Since the CIFAR-10 dataset does not provide a transition matrix, we need to design a transition matrix estimation algorithm. In this study, we used the anchor point assumption method and the T-Revision method based on the previously implemented MLP and ResNet-18 models to estimate the transition matrices for FashionMINIST0.3 and FashionMINIST0.6. By calculating the error with respect to the provided true transition matrices, we identified the best transition matrix estimation method to estimate transition matrix of CIFAR-10.
\subsection{Estimate T with Archor Point Assumption}
The Anchor Point Assumption method is a technique used to estimate the transition matrix \( T \). This approach assumes that for each class, there exists at least one "anchor point," a sample that is highly likely to belong to a specific class with minimal or no noise. Using these anchor points, we can estimate the transition matrix \( T \), which represents the probability of a true label being observed as a noisy label.

One way to apply this method is to assume that there are anchor points within the training samples and to treat instances with a high posterior probability for the \(i\)-th noisy class as anchor points for the \(i\)-th clean class. The probability distribution for this anchor point will then represent the \(i\)-th row of the transition matrix. Thus, for each label, we select samples with the highest predicted probabilities for that label. Using these samples, we can estimate the transition matrix by taking the predicted probabilities of all labels for these anchor points, assuming the anchor point hypothesis holds.

In practice, if we directly select the sample with the highest probability in each class as the anchor point, issues such as model overfitting risk and class imbalance may cause the probability of this sample to be overly high. Therefore, existing studies typically select samples within the 97th percentile of the probability distribution in each class as anchor points. This approach effectively reduces the impact of extreme sample errors, resulting in a more robust estimate of the transition matrix. Inspired by the law of large numbers, this paper further refines the transition matrix by averaging multiple estimations of each element, yielding a more accurate transition matrix.

\subsection{Estimate T with T-Revision Method}
In practical applications, estimating the transition matrix \( T \) is not always accurate. To learn the label transition matrix, we need to estimate the conditional probability \( P(\bar{Y} | X = x) \). However, this estimation process may introduce errors. For example, if we take a sample \( x_i \) as an anchor point for the \( i \)-th class, but the actual probability \( P(Y = i | X = x_i) \neq 1 \), meaning that this sample does not fully represent the \( i \)-th class, then the resulting transition matrix will no longer be a standard identity matrix but rather a non-identity matrix \( L \). Consequently, the learned transition matrix becomes \( T L \) , leading to cumulative errors. Based on such inaccurate transition matrices, the performance of existing algorithms may degrade significantly. \cite{xia2019anchor}

Based on this, T-Reversion combines the ideas of forward correction and importance reweighting, proposing a framework that simultaneously estimates both the true \( T \) and \( P(Y | X) \).

The proposed training process consists of three steps. First, we minimize the cross-entropy loss to train a classifier. This classifier is then used to estimate an initial transition matrix, $\hat{T}$. The estimation of $\hat{T}$ is based on the samples with the highest predicted posterior probability $P(\bar{Y} \mid X = x)$, assuming that the most confident samples are closer to the clean labels.

Next, using the previously estimated transition matrix $\hat{T}$, we train a new model by minimizing a weighted loss function as mentioned in \ref{eq:loss}. 

In the final step, we introduce a slack variable $\Delta T$, which modifies the initial transition matrix $T$. The slack variable $\Delta T$ serves to refine $T$ and improve its adaptability to noisy data. According to the above description, the original correction formula for the transition matrix is:
\begin{equation}
T = \hat{T} + \Delta T,
\end{equation}

By jointly training the classifier and $\Delta T$, we ensure that both the transition matrix and the classifier are optimized simultaneously. The mathematical modification is mentioned in the calculation of the weight \( \beta \) for each sample in the weighted loss function.

 \begin{equation}
    \beta = \frac{P(\bar{Y} \mid X = x)}{\hat{P}(\bar{Y} \mid X = x)} = \frac{g_{\bar{Y}}(x)}{((\hat{T} + \Delta T)^\top g(x))_{\bar{Y}}},
\end{equation}
This joint training enables the model to better capture the distribution of noisy labels, ensuring a more robust performance in handling noisy data. Ultimately, the classifier and the refined transition matrix work together to estimate both posterior distributions, leading to improved overall performance in noisy environments.

However, during training, we observed that this approach led to negative values in the loss function, and the convergence process was unstable. Therefore, we introduced two main improvement strategies in this paper.

\subsubsection{T-Revision Softmax Method}
First, to ensure that the transition matrix \( T \) satisfies the probability property (i.e., each row sums to 1 and elements are non-negative), we applied the softmax function to \( \hat{T} + \Delta T \). Directly using \( \hat{T} + \Delta T \) may result in negative values or rows that do not sum to 1, while the softmax function normalizes each row to represent a valid probability distribution. This leads to:
\begin{equation}
T = \text{softmax}(\hat{T} + \Delta T),
\end{equation}
where the softmax function is defined as:
\begin{equation}
\text{softmax}(z_i) = \frac{\exp(z_i)}{\sum_j \exp(z_j)},
\end{equation}
where \(z_i\) represents the \(i\)-th element of the input vector \(z\), \(\exp(z_i)\) is the exponential of \(z_i\), \(\sum_j \exp(z_j)\) is the sum of exponentials of all elements in \(z\), and \(\text{softmax}(z_i)\) represents the probability of \(z_i\) in the resulting probability distribution, where the sum of all probabilities is 1.

Although this method ensures stable improvement of classifier performance during training, applying softmax to the \( T \) matrix disrupts the original probability distribution, resulting in a higher Relative Reconstruction Error (RRE) for the corrected \( T \) matrix compared to before correction.
\subsubsection{T-Revision Alpha Method}
Therefore, we introduced a more transition-matrix-friendly improvement. We added an adjustment parameter \( \alpha \) to the original formula to control the influence of \( \Delta T \) on the transition matrix. When the noise level is high or the transition matrix is highly asymmetric, a larger \( \alpha \) can be chosen to enhance the correction of \( T \); conversely, a smaller \( \alpha \) can be chosen to retain more of the initial estimate. The improved formula is as follows:
\begin{equation}
T = \hat{T} + (\alpha \cdot \Delta T),
\end{equation}
In this paper, the adjustment parameter is set to \( \alpha = 0.01 \).ReLU (Rectified Linear Unit) is a commonly used activation function defined as:
\begin{equation}
\text{ReLU}(x) = \max(0, x),
\end{equation}
This function outputs \( x \) if \( x \geq 0 \) and 0 otherwise.

In the context of adjusting the transition matrix, we apply ReLU to ensure that all elements of \( T \) are non-negative:
\begin{equation}
T = \text{ReLU}(T),
\end{equation}

This step is necessary because a transition matrix should ideally represent probabilities, which cannot be negative. 

\subsection{Experiments}
\subsubsection{Evaluation Metrics}

The Relative Reconstruction Error (RRE) is used here to assess the accuracy of transition matrix estimation by evaluating how closely the estimated matrix aligns with the true transition matrix. RRE, being scale-invariant, normalizes the difference between the true and estimated matrices, providing a relative measure of accuracy that highlights proportionate differences rather than absolute errors. This makes it particularly effective for transition matrix estimation, as it emphasizes the overall fidelity of the estimation without being overly influenced by large discrepancies of individual element that might distort Mean Squared Error (MSE) results.Given two matrices \( A \) and \( B \), the Relative Reconstruction Error (RRE) is calculated as follows:

\begin{equation}
\text{RRE}(A, B) = \frac{\| A - B \|_F}{\| A \|_F},
\label{eq:RRE}
\end{equation}

where:
\begin{itemize}
    \item \( \| \cdot \|_F \) denotes the Frobenius norm of a matrix.
    \item \( A - B \) represents the element-wise difference between the matrices.
\end{itemize}

The Frobenius norm is defined as:

\begin{equation}
\| A \|_F = \sqrt{\sum_{i=1}^{m} \sum_{j=1}^{n} |a_{ij}|^2},
\label{eq:FrobeniusNorm}
\end{equation}

Substituting this into the RRE formula, we get:

\begin{equation}
\text{RRE}(A, B) = \frac{\sqrt{\sum_{i=1}^{m} \sum_{j=1}^{n} |a_{ij} - b_{ij}|^2}}{\sqrt{\sum_{i=1}^{m} \sum_{j=1}^{n} |a_{ij}|^2}},
\label{eq:RREExpanded}
\end{equation}

Equation \eqref{eq:RRE} calculates the RRE between the original matrix \( A \) and the reconstructed matrix \( B \). A lower RRE indicates that \( B \) closely approximates \( A \).

\subsubsection{Experiement Setting}
\label{5.3.2}
To ensure rigorous performance evaluation, we conducted 10 training runs on each noise transition matrix estimation model on the FashionMINIST0.3, FashionMINIST0.6, and CIFAR-10 datasets. For the FashionMINIST0.3 and FashionMINIST0.6 datasets, which have known true transition matrices, we calculated the error of the estimated transition matrix obtained from each training run for each method and further computed the mean and standard deviation to comprehensively assess performance and stability.

We then computed a new transition matrix by averaging each element of the 10 estimated transition matrices and compared the estimation error of this averaged matrix with the mean estimation error of the individual 10 matrices.

Notably, the default parameters for T-revision series methods differ from other methods during training. T-revision primarily involves fine-tuning the model, and according to the authors’ code implement \cite{xia2019anchor}, T-revision uses a very small learning rate of 0.0000005 and a relatively large batch size (128). We maintained the designed learning rate and increased the batch size to 256 based on practical considerations. Other methods used a learning rate of 0.0005 and a batch size of 32, which are results from the parameter selection experiment discussed in Chapter 3 \ref{3}.

\subsubsection{Experiments Results and Discussions}
We estimate the transition matrix on both the FasionMINIST0.3 and FasionMINIST0.6 dataset and compare the estimated transition matrix with the true transition matrix to demonstrate the effective of our modified T-Revision method. Since the loss of the standard T-revision method is negative, and not able to be properly trained, we only used the modified version T-Revision-Softmax and the T-Revision-Alpha method to predict the labels and obtain an estimated transition matrix.

Table \ref{tb:matrix 0.3} shows the results of different transition matrix estimators. RRE Mean represents the average Relative Reconstruction Error (RRE) across ten experiments, while Mean Matrix RRE calculates the RRE of the mean matrix obtained from these ten experiments. The matrix \( T_{Estimated} \) is estimated using the method with the minimum Mean Matrix RRE, which is ResNet T-Revision Alpha here.

\begin{minipage}{0.45\textwidth}
\[
T_{\text{Estimated}} = \begin{bmatrix}
0.7005 & 0.2867 & 0.0058 & 0.0069 \\
0.0015 & 0.6970 & 0.2991 & 0.0023 \\
0.0054 & 0 & 0.7392 & 0.2569 \\
0.2692 & 0.0016 & 0.0105 & 0.7187 \\
\end{bmatrix}
\]
\end{minipage}
\hfill
\begin{minipage}{0.45\textwidth}
\[
T_{\text{True}} = \begin{bmatrix}
0.7 & 0.3 & 0 & 0 \\
0 & 0.7 & 0.3 & 0 \\
0 & 0 & 0.7 & 0.3 \\
0.3 & 0 & 0 & 0.7 \\
\end{bmatrix}
\]
\end{minipage}

\begin{table}[H]
\centering

\begin{tabular}{lccc}
\hline
\textbf{Dataset} & \textbf{RRE Mean} & \textbf{RRE STD} & \textbf{Mean Matrix RRE} \\
\hline
ResNet Anchor Point & 0.1033 & 0.0001 & 0.0475 \\
ResNet T-Revision Softmax & 0.589 & $2.36 \times 10^{-13}$ & 0.589 \\
ResNet T-Revision Alpha & \textbf{0.0469} & \textbf{$5.18 \times 10^{-14}$} & \textbf{0.0469} \\
MLP Anchor Point & 0.0987 & 0.001 & 0.0634 \\
MLP T-Revision Softmax & 0.509 & $6.04 \times 10^{-11}$ & 0.509 \\
MLP T-Revision Alpha & 0.0621 & $6.19 \times 10^{-14}$ & 0.0621 \\
\hline
\end{tabular}
\caption{Matrix Comparison Result for FashionMINIST0.3}
\label{tb:matrix 0.3}
\end{table}

In terms of both RRE Mean and Mean Matrix RRE, ResNet T-Revision Alpha and MLP T-Revision Alpha performed the best, achieving the lowest error values, which demonstrates the superior performance and high accuracy of T-Revision Alpha in estimating the transition matrix. In contrast, although ResNet T-Revision Softmax and MLP T-Revision Softmax have relatively low RRE STD, their RRE Mean and Mean Matrix RRE values are comparatively high, indicating lower accuracy for this method. Meanwhile, the anchor point model achieved decent RRE Mean and Mean Matrix RRE, but its fluctuations were the largest, reflecting instability in the method.

Table \ref{tb:matrix 0.6} shows the results for matrix estimators on the FashionMNIST0.6 dataset. Similar to the previous dataset, T-Revision still achieves the lowest RRE Mean and Mean Matrix RRE. However, this time, the MLP method achieves the lowest value of 0.0088. The experiments on both datasets demonstrate the effectiveness of the T-Revision Alpha method in predicting the transition matrix.

We can see that there is almost no difference between the average error of the transfer matrices estimated by the model obtained for each training of the T-Revision series and the error of the average matrices of these transfer matrices, which is due to the fact that T-Revision is mainly fine-tuning the model so the learning rate is very small, and also that the initial matrices have already been estimated well enough, leading to a small variance of the multiple revisions, which is also reflected in the table .

\begin{minipage}{0.45\textwidth}
\[
T_{\text{Estimated}} = \begin{bmatrix}
0.3967 & 0.2016 & 0.2038 & 0.1978 \\
0.2021 & 0.3967 & 0.2001 & 0.2011 \\
0.1970 & 0.2006 & 0.3995 & 0.2029 \\
0.1962 & 0.2008 & 0.2016 & 0.4014 \\
\end{bmatrix}
\]
\end{minipage}
\hfill
\begin{minipage}{0.45\textwidth}
\[
T_{\text{True}} = \begin{bmatrix}
0.4 & 0.2 & 0.2 & 0.2 \\
0.2 & 0.4 & 0.2 & 0.2 \\
0.2 & 0.2 & 0.4 & 0.2 \\
0.2 & 0.2 & 0.2 & 0.4 \\
\end{bmatrix}
\]
\end{minipage}

\begin{table}[H]
\centering
\begin{tabular}{lccc}
\hline
\textbf{Dataset} & \textbf{RRE Mean} & \textbf{RRE STD} & \textbf{Mean Matrix RRE} \\
\hline
ResNet Anchor Point & 0.1342 & 0.0017 & 0.0318 \\
ResNet T-Revision Softmax & 0.4344 & $5.54 \times 10^{-12}$ & 0.4344 \\
ResNet T-Revision Alpha & 0.0301 & \textbf{$2.54 \times 10^{-16}$} & 0.0301 \\
MLP Anchor Point & 0.1117 & 0.0007 & 0.0581 \\
MLP T-Revision Softmax & 0.3837 & $5.64 \times 10^{-11}$ & 0.3837 \\
MLP T-Revision Alpha & \textbf{0.0088} & $7.40 \times 10^{-15}$ & \textbf{0.0088} \\
\hline
\end{tabular}
\caption{Matrix Comparison Result for FashionMINIST0.6}
\label{tb:matrix 0.6}
\end{table}

We also test the performance of the estimated transition matrix on the two dataset, and the results are shown in Table \ref{tab:MINIST0.3 RESULT 1} and Table \ref{tab:MINIST0.6 RESULT 1}. 
The T-Reversion-Alpha method achieves the best performance among all models in small noise rate situation, with the test loss reduced to 0.1211 and accuracy increased to 96.3375\%, along with the smallest standard deviations (0.0022 and 0.0527). 
In large noise rate(0.6) situation, while T-Revision-alpha has minimal loss, T-Revision-Softmax has higher accuracy.
Although T-Reversion-Alpha outperforms the T-Revision-Softmax, the T-Revision-Softmax still outperforms the forward correcting and importance reweighting methods \ref{tab:MINIST0.3 RESULT} \ref{tab:MINIST0.6 RESULT}, demonstrating the excellent robustness and stability of our improved model in handling label noise. The same pattern can also be observed in MLP models.

\begin{table}[h]
    \centering
    \begin{tabular}{lcccc}
        \toprule
        \textbf{Method} & \textbf{Loss Mean} & \textbf{Loss STD} & \textbf{Accuracy Mean (\%)} & \textbf{Accuracy STD} \\
        \midrule
        T-Revision-Softmax ResNet & 0.1326 & 0.0032 & 95.9325 & 0.1067 \\
        T-Revision-Alpha ResNet & \textbf{0.1211} & \textbf{0.0022} & \textbf{96.3375} & \textbf{0.0527} \\
\hline
        T-Revision-Softmax MLP & 0.2112 & \textbf{0.0041} & \textbf{93.7950} & \textbf{0.1182} \\
        T-Revision-Alpha MLP & \textbf{0.2045} & 0.0042 & 93.5900 & 0.1295 \\
        \bottomrule
    \end{tabular}
\caption{Testset Performance Metrics for Different Models on FashionMINIST0.3}
\label{tab:MINIST0.3 RESULT 1}
\end{table}

\begin{table}[h]
    \centering
    \begin{tabular}{lcccc}
        \toprule
        \textbf{Method} & \textbf{Loss Mean} & \textbf{Loss STD} & \textbf{Accuracy Mean (\%)} & \textbf{Accuracy STD} \\
        \midrule
        T-Revision-Softmax ResNet & 0.3138 & 0.0181 & \textbf{91.7550} & 0.2043 \\
        T-Revision-Alpha ResNet & \textbf{0.2966} & \textbf{0.0138} & 91.5375 & \textbf{0.1085} \\
        \hline
        T-Revision-Softmax MLP & 0.3261 & 0.0054 & \textbf{91.6275} & \textbf{0.0825} \\
        T-Revision-Alpha MLP & \textbf{0.2956} & \textbf{0.0025} & 91.3000 & 0.1792 \\
        \bottomrule
    \end{tabular}
    \caption{Testset Performance Metrics for Different Models on FashionMINIST0.6}
    \label{tab:MINIST0.6 RESULT 1}
\end{table}

In addition, we visualised the accuracy and Loss of the models trained on the above two datasets on the training and test sets in the form of box plots, in order to save space we will put them in the appendix \ref{fig:mnist3} \ref{fig:mnist6}, and no longer analyse them specifically, the higher the average represents the better the model is, the longer the box is or the appearance of circles representing outliers represents that the model is not stable enough.

\section{Label Noise Methods With Unknown Flip Rates}
\label{unknown noise method}
\subsection{Dataset Description and Preprocessing}
We use the CIFAR-10 dataset provided by USYD teaching team to explore the performance of our method. The dataset contains 20000 tarining and validation samples, while the number of test samples is 4000. Each example is a colorful image of \(32\times32\) with 3 channels. The class set of labels of all examples is \({0, 1, 2, 3}\). The training set labels have label noises, while the test set does not involve them. For this dataset, the preprocessing process includes image resizing, color channel adjustment, and conversion to tensors. First, using transforms, we ensure all images are set to a fixed size of 32×32 to match the required input size for ResNet18 (CIFAR-10) and ensure the channel order is in the standard three-channel RGB format. Then, we convert the images to tensor format to meet PyTorch requirements. When training the MLP model, we flatten the image tensor from three dimensions (channel, height, and width) into a single-dimensional tensor. As the same as FachionMINIST dataset, we chose not to normalize the pixel value range, considering that we did not use any pre-trained models,.

\subsection{Experiement}

\subsubsection{Experiement Setting}
In this dataset, we will estimate the initial transition matrix based on the anchor point assumption method and apply this transition matrix to the forward correction and importance reweighting methods, then compare their results. Subsequently, we will use the model obtained from the importance reweighting method and the initial transition matrix as inputs to the T-Revision series methods, and apply two modified T-Revision methods to correct the initial transition matrix to obtain the final transition matrix and the clean label classification model.

The hyperparameters for model training are consistent with those mentioned in the previous chapter \ref{5.3.2}. For each method, we will conduct ten experiments, averaging the test loss and accuracy to reduce randomness, and calculating the standard deviation to demonstrate stability.

\subsubsection{Experiments Results and Discussions}
The results for CIFAR-10 datasets in shown in Table \ref{tab:CIFAR10 results}.

\begin{table}[h]
    \centering

    \begin{tabular}{lcccc}
        \toprule
        \textbf{Method} & \textbf{Loss Mean} & \textbf{Loss STD} & \textbf{Accuracy Mean (\%)} & \textbf{Accuracy STD} \\
        \midrule
        Baseline ResNet & 0.5115 & 0.0238 & 83.5575 & 0.7038 \\
        Forward ResNet & 0.4250 & 0.0349 & 84.1550 & 0.8966 \\
        Importance ResNet & 0.4394 & 0.0333 & 84.2300 & 0.6524 \\
        T-Revision-Softmax ResNet & \textbf{0.4037} & 0.0070 & \textbf{86.1550} & 0.2750 \\
        T-Revision-Alpha ResNet & 0.4068 & \textbf{0.0045} & 84.9875 & \textbf{0.2478} \\
        \hline
        Baseline MLP & 0.9514 & 0.0503 & 65.3200 & 1.8173 \\
        Forward MLP & 0.9694 & 0.1248 & 67.6125 & 0.8753 \\
        Importance MLP & 0.9765 & 0.1130 & 67.3950 & 0.5802 \\
        T-Revision-Softmax MLP & \textbf{0.8223} & \textbf{0.0062} & \textbf{68.3575} & 0.2245 \\
        T-Revision-Alpha MLP & 0.9003 & 0.0143 & 68.0125 & \textbf{0.1689} \\
        \bottomrule
    \end{tabular}
    \caption{Testset Performance Metrics for Different Models on CIFAR-10}
    \label{tab:CIFAR10 results}
\end{table}

Based on the results in Table \ref{tab:CIFAR10 results}, we can observe that ResNet and MLP models with different methods display significant differences in test loss, accuracy, and standard deviation on the CIFAR-10 dataset. Overall, the T-Revision series methods demonstrate superior performance across all metrics, highlighting their advantage in handling label noise.

For the ResNet model, the baseline model’s mean test loss is 0.5115, with an accuracy of 83.5575\%, and relatively high standard deviations for both loss and accuracy, indicating limited robustness when label noise is present. After applying the Forward Correction method, ResNet's accuracy improves slightly to 84.155\% with a slight reduction in test loss, though standard deviation changes remain minimal. This suggests that Forward Correction mitigates the impact of label noise to some extent, though the improvement is limited. The Importance Reweighting method further raises the accuracy to 84.23\% while slightly reducing the accuracy variance, showing some improvement. However, T-Revision-Softmax ResNet achieves the best results among all methods, with a test loss reduced to 0.4037 and accuracy increased to 86.155\%. The standard deviations for loss and accuracy, at 0.0070 and 0.2750 respectively, indicate excellent stability and robustness. Although the T-Revision-Alpha ResNet achieves a slightly lower accuracy of 84.9875\%, it has the lowest loss standard deviation of 0.0045, underscoring T-Revision-Alpha’s strong stability.

A similar trend is observed in the MLP model. The baseline MLP model’s mean test loss is 0.9514, with an accuracy of 65.32\% and high standard deviations for both loss and accuracy, indicating instability in the presence of noisy labels. The Forward Correction method improves MLP's accuracy to 67.6125\%, but changes in mean loss and standard deviation remain minimal, suggesting that its effect is less pronounced in MLP than in ResNet. The Importance Reweighting method performs similarly, raising accuracy to 67.3950 and slightly reducing both loss and accuracy variance. However, T-Revision-Softmax MLP demonstrates the best performance among all MLP methods, with test loss decreased to 0.8223 and accuracy increased to 68.3575\%. The significant reduction in loss and accuracy standard deviations, at 0.0062 and 0.2245 respectively, indicates greatly enhanced stability and performance. T-Revision-Alpha MLP achieves a slightly lower accuracy of 68.0125 but with lower loss and standard deviations, showcasing excellent stability.

In summary, the T-Revision series methods, particularly T-Revision-Softmax, exhibit significant superiority in both ResNet and MLP structures. Through a combination of correction and reweighting, the T-Revision series effectively counters the impact of label noise, enhancing the robustness and stability of the model. In comparison, the Forward Correction and Importance Reweighting methods provide limited improvements and are less effective than the T-Revision series, especially in the MLP model.

As in the previous chapter, we also visualise the accuracy and Loss of the model trained on the CIFAR-10 dataset on both the training and test sets, which we also put in the Appendix \ref{fig:cifar10}.

Although T-Revision-Softmax achieves the highest score in the prediction task, it destroys the original probability distribution of the transfer matrix, so we preferred the transfer matrix estimated by T-Revision-Alpha, which has the second highest accuracy, as the final transfer matrix estimation result.
Below is the transfer matrix obtained by using ResNet as a classifier and the T-Revision-alpha method as a transfer matrix estimator:
\begin{equation}
    T_{Estimate} = \begin{bmatrix}
0.8858 & 0.0962 & 0.0078 & 0.0102 \\
0.0058 & 0.9053 & 0.0900 & 0 \\
0.0030 & 0 & 0.9219 & 0.0786 \\
0.0906 & 0 & 0.0247 & 0.8854 \\
\end{bmatrix}
\end{equation}

\section{Conclusion}
In this study, we independently implemented two basic classifiers, MLP and ResNet18, along with three loss functions that are robust to label noise: forward correction, importance reweighting, and the T-reversion method. To address the issue of negative losses observed in T-reversion, we introduced two improved strategies: T-reversion-Alpha and T-reversion-Softmax. Additionally, we provided the anchor point assumption method and T-reversion for transition matrix estimation.

For the FashionMINIST0.3 and FashionMINIST0.6 datasets, which have known noise transition matrices, we used the provided noise matrices to train the MLP and ResNet18 models and tested the robustness of the three loss functions. Next, we estimated the transition matrices using the two estimation methods mentioned earlier, calculated the Relative Residual Error (RRE) between the estimated and true transition matrices to evaluate the noise estimation performance, and selected T-reversion-Alpha based on ResNet18 as the optimal transition matrix estimation method due to its lower RRE and higher accuracy on the test set.

For the CIFAR-10 dataset, where the transition matrix is unknown, we applied the optimal transition matrix estimation method identified earlier to derive the noise matrix for CIFAR-10. Using this matrix, we tested the performance of the label-noise-robust loss functions on the testset for both MLP and ResNet18 models. 

\subsection{Meaningful conclusions and Findings}
Our main contributions are as follows:

\begin{enumerate}
    \item We address the issue of non-robust training in the T-Revision method by proposing two improved approaches: the \textbf{T-Revision-Alpha} method, which enables more accurate estimation of the transition matrix, and the \textbf{T-Revision-Softmax} method, which provides a more precise output of clean label classification probabilities. Comparative experiments demonstrate that T-Revision-Alpha yields the lowest estimation error for the transition matrix in terms of RRE, while T-Revision-Softmax achieves the best performance in terms of average test accuracy and the mean and standard deviation of test loss.

    \item Inspired by the law of large numbers, we improve the transition matrix estimation by averaging multiple estimations of each element in the matrix, resulting in a more accurate transition matrix. Experiments show that this approach can reduce the estimation error by an order of magnitude in anchor point methods, and also provides a stable, small reduction in error for the T-Revision-Alpha method.
\end{enumerate}

The primary findings of our work are as follows:

\begin{enumerate}
    \item Based on the mean and standard deviation of test accuracy and test loss, we observe that for all datasets and classifiers, the forward correction method generally outperforms importance re-weighting methods when transition matrix correction is not considered. This may be due to the reliance of importance re-weighting methods on the assumption that the target conditional distribution follows a link function (Sigmoid).

    \item When considering transition matrix correction, the T-Revision-Softmax method generally outperforms other methods, except for the case of T-Revision-Alpha on the ResNet trained with FashionMINIST0.3, where it performs better in terms of test accuracy. In terms of the mean and standard deviation of the loss function, T-Revision-Alpha consistently performs better across all classifiers for the FashionMINIST grayscale datasets, while T-Revision-Softmax is more effective for the CIFAR-10 dataset with more complex sample features. Both methods significantly outperform others, validating the effectiveness of our work.

    \item From the test set accuracy, it can be seen that the label noise robust loss function does not significantly improve model performance. However, based on the cross-entropy on the test set, the label noise robust loss function has a notable effect in reducing cross-entropy.

    \item Although the CIFAR-10 dataset has the lowest noise rate, its three-channel images yield poorer prediction results compared to simpler datasets. This suggests that the impact of sample feature complexity on model performance is greater than that of noise rate. However, within the same dataset, a lower noise rate corresponds to higher model accuracy and lower loss.

    \item Regarding transition matrix estimation, FashionMINIST0.3 with a lower noise rate actually performs worse than FashionMINIST0.6 with a higher noise rate. We find that although the former has a lower noise rate, it lacks symmetry, whereas the noise matrix of the latter is fully symmetric. We hypothesize that the effectiveness of transition matrix estimation is more closely related to the symmetry of the true noise matrix than to the noise rate itself; that is, symmetry has a greater influence on estimation error than noise rate.
\end{enumerate}

\subsection{Future Work}

Our future work includes the following:

\begin{enumerate}
    \item We will experimentally verify our hypothesis by controlling for identical noise rates with varying matrix symmetries, to examine the impact of symmetry on transition matrix estimation error. Additionally, we will control matrix symmetry and test the impact of different levels of label noise on estimation error. Such comparative experiments are missing from the existing literature.

    \item T-Revision currently requires training three times: first, to estimate the T matrix; second, to train a noise-robust classifier for fine-tuning; and finally, to correct the T matrix and fine-tune the classifier. This approach is more robust when noise rates are high and matrix symmetry is poor. However, such extreme conditions rarely appear in practical classification tasks. Therefore, we aim to initialize with an identity matrix or a symmetric matrix with a low noise rate, and then refine this initialized matrix. For the correction matrix, a custom loss function should be designed to ensure row sums approach zero. A dynamic adjustment mechanism will be implemented to balance the influence of this loss function and the classifier's loss on matrix parameter optimization.

    \item We plan to test our methods on more challenging tasks involving label noise learning, such as object detection, expert models, semi-supervised learning, and knowledge distillation.
\end{enumerate}

\newpage
\bibliographystyle{unsrt}
\bibliography{ref}
\newpage
\section*{Appendix}
\subsection*{Code Guidance}
\label{A}
The code link of this report is below: 
\href{https://colab.research.google.com/drive/1f0SGvbRBNp4Jy1TOy6aFe38tcqnk3mKz?usp=sharing}{Colab Notebook Link}

After opening the link, please change the runtime type to 'A100'. 

\begin{figure}[H]
    \centering
    \begin{subfigure}[b]{0.49\textwidth}
        \centering
        \includegraphics[width=\textwidth]{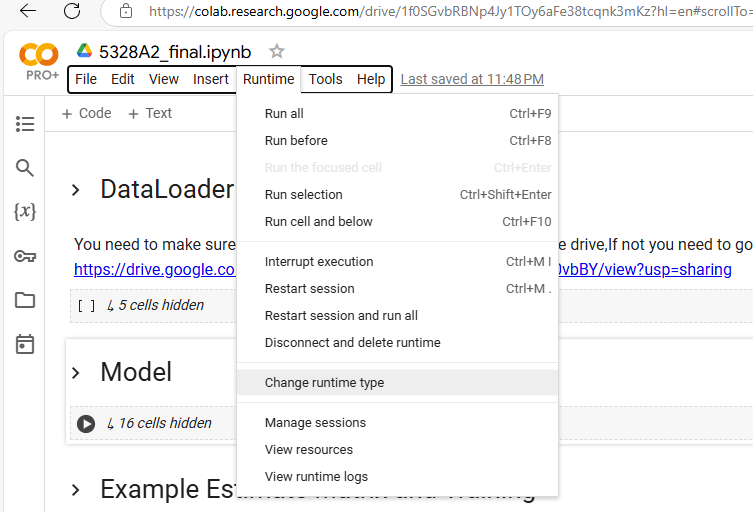}
        \caption{Click Runtime type}
        \label{Click Runtime type}
    \end{subfigure}
    \hfill
    \begin{subfigure}[b]{0.49\textwidth}
        \centering
        \includegraphics[width=\textwidth]{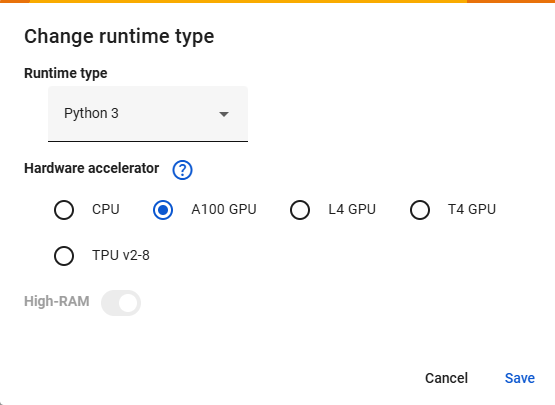}
        \caption{Example of White Block Noise 2.}
        \label{Choose CPU High RAM}
    \end{subfigure}
    \caption{Choose CPU High RAM}
    \label{Choose config}
\end{figure}

Then, start running the code from the first block sequentially. If the data download fails, you can manually download the data from the link provided in the notebook and upload it to the Content directory.

The dataset can be downloaded from the following link: 
\href{https://drive.google.com/file/d/1IAaf3yDa1KaRXNmphLjnb9LJKMF0vbBY/view?usp=sharing}{Download Dataset}

The experimental results table can be downloaded from the following link: 
\href{https://drive.google.com/file/d/19_ui2U6TvpzguuhgmOJegvJKDJX0LnYS/view?usp=sharing}{Download Experimental Results Table}

If you only want to check the experimental results rather than verifying if the experiment code runs successfully, please skip the "Experiments" Block and directly run the "Plots" Block. Running the full experiment takes a long time, and the results have already been collected in a CSV archive, which are visualized in the "Plots" Block.

\begin{figure}[H]  % htbp 是图片的插入位置参数
    \centering
    \includegraphics[width=1\textwidth]{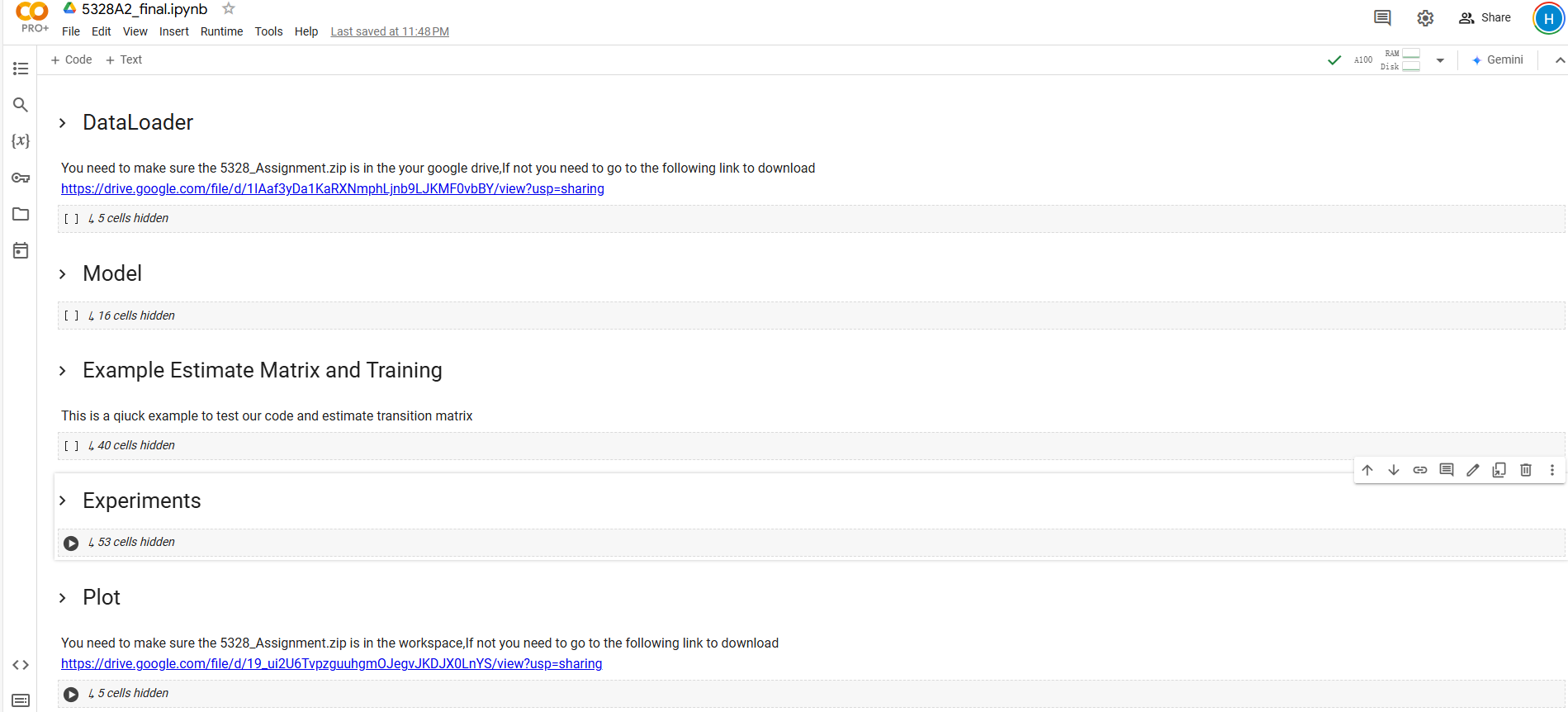}
    \caption{Example of White Block Noise.}
    \label{Structure of Code}
\end{figure}

\subsection*{Plots}
\begin{figure}[H]
    \centering
    \includegraphics[width=1\textwidth]{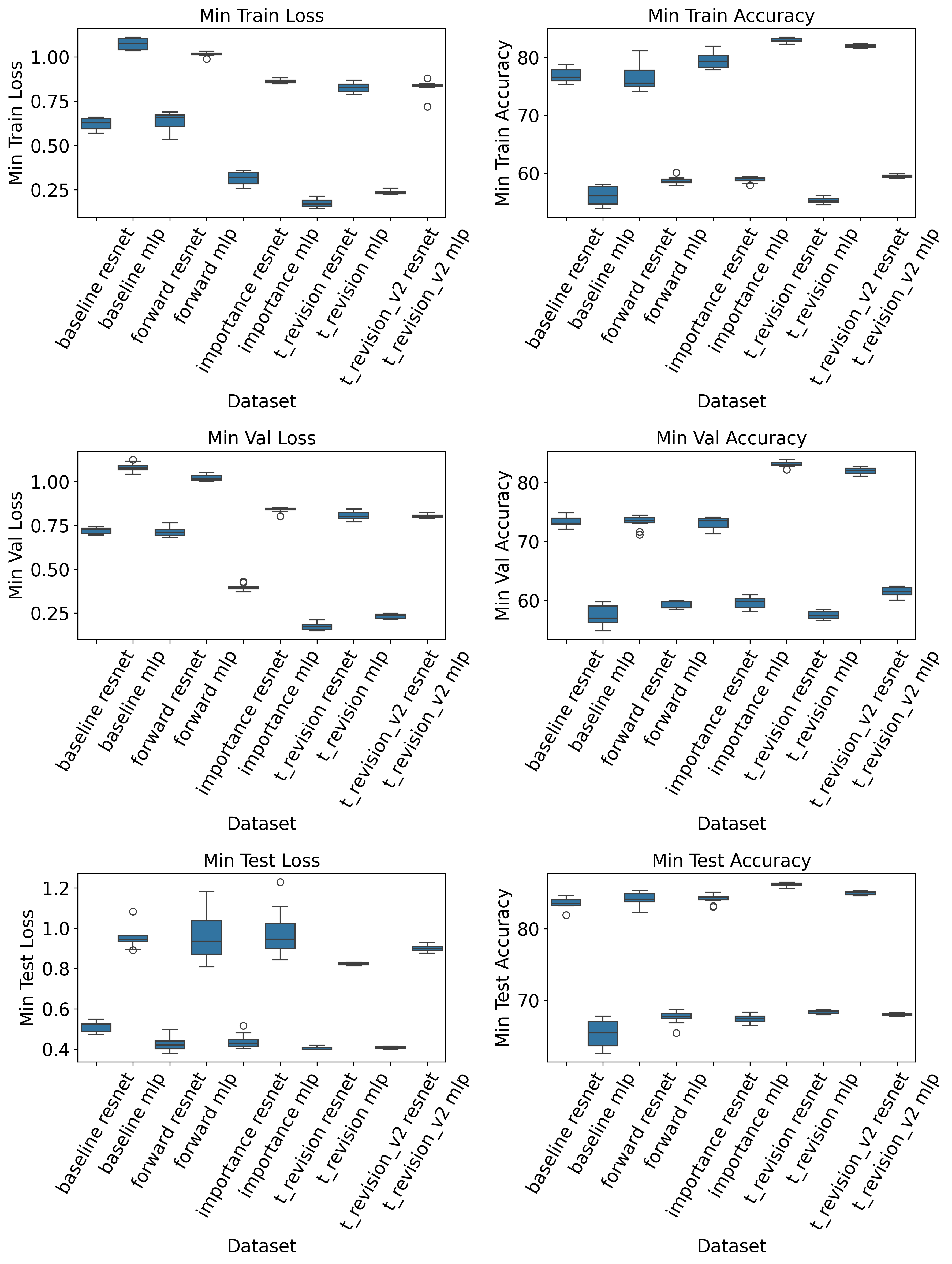}
    \caption{Boxplots of CIFAR-10}
    \label{fig:cifar10}
\end{figure}

\newpage % 手动分页

\begin{figure}[H]
    \centering
    \includegraphics[width=1\textwidth]{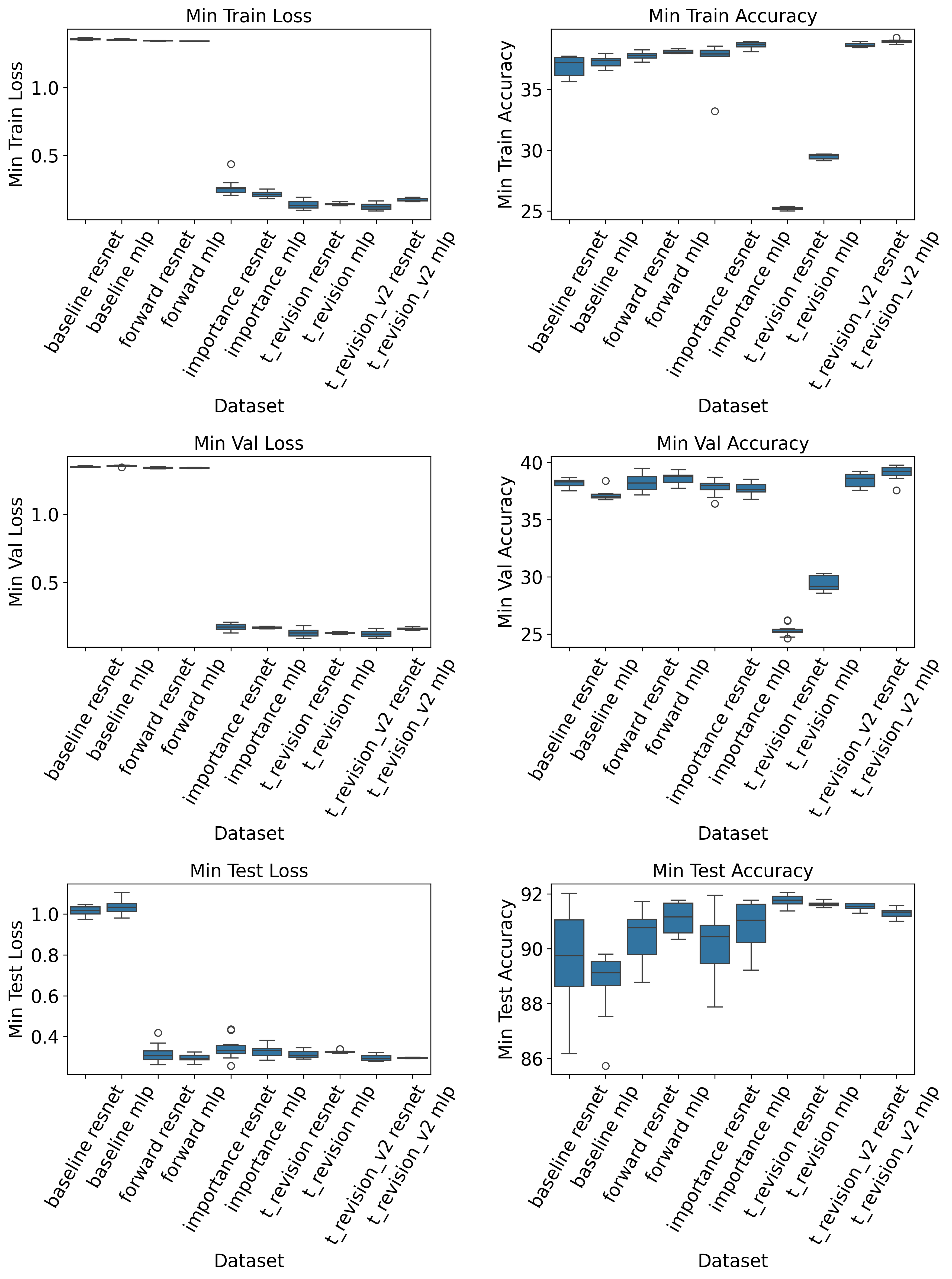}
    \caption{Boxplots of FashionMINIST0.6}
    \label{fig:mnist6}
\end{figure}

\newpage % 手动分页

\begin{figure}[H]
    \centering
    \includegraphics[width=1\textwidth]{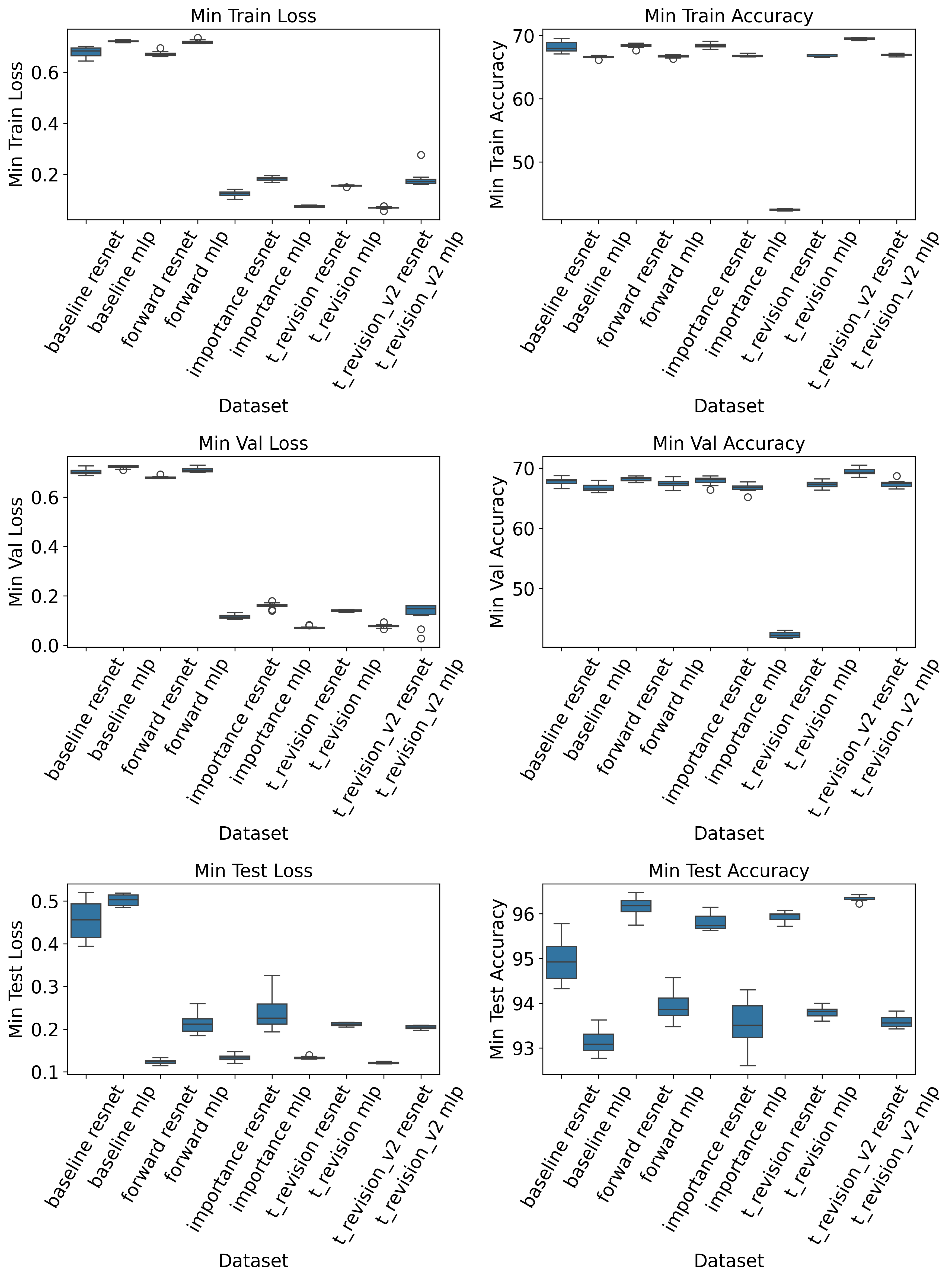}
    \caption{Boxplots of FashionMINIST0.3}
    \label{fig:mnist3}
\end{figure}

\end{document}